\documentclass[reprint,aps,pre]{revtex4-1}

\usepackage{amsmath,amsfonts,amscd,amssymb,graphicx}

\usepackage{dcolumn}
\usepackage{bm}
\usepackage{overpic}
\usepackage{color}
\usepackage{adjustbox}
\usepackage[caption=false]{subfig}
\usepackage{rotating}
\usepackage{float}
\makeatletter
\let\newfloat\newfloat@ltx
\makeatother
\usepackage{algorithm}
\usepackage{hyperref}
\usepackage{algpseudocode}
\algnewcommand\algorithmicinput{\textbf{Input:}}
\algnewcommand\Input{\item[\algorithmicinput]}

\begin{document}

\title{Koopman-theoretic Approach for Identification of Exogenous Anomalies\\in Nonstationary Time-series Data}

\author{Alex Mallen$^{*,\dag}$, Christoph A. Keller$^{**,\ddag}$, and J. Nathan Kutz$^\dag$}
 \affiliation{$^*$Department of Computer Science, University of Washington, Seattle, WA 98195}
 \affiliation{$^{**}$NASA Global Modeling and Assimilation Office, Goddard Space Flight Center, Greenbelt, MD 20771}
 \affiliation{$^\ddag$Morgan State University, Baltimore, MD 21251} 
 \affiliation{$^\dag$Department of Applied Mathematics and Electrical and Computer Engineering, University of Washington, Seattle, WA 98195} 

\begin{abstract}
In many scenarios, it is necessary to monitor a complex system via a time-series of observations and determine when anomalous exogenous events have occurred so that relevant actions can be taken.  Determining whether current observations are abnormal is challenging. It requires learning an extrapolative probabilistic model of the dynamics from historical data, and using a limited number of current observations to make a classification. We leverage recent advances in long-term probabilistic forecasting, namely {\em Deep Probabilistic Koopman}, to build a general method for classifying anomalies in multi-dimensional time-series data. We also show how to utilize models with domain knowledge of the dynamics to reduce type I and type II error. We demonstrate our proposed method on the important real-world task of global atmospheric pollution monitoring, integrating it with NASA's Global Earth System Model. The system successfully detects localized anomalies in air quality due to events such as COVID-19 lockdowns and wildfires.
\end{abstract}
\maketitle

\section{INTRODUCTION} \label{intro}

{\em Time-series analysis} is used to extracting meaningful statistics and characteristics of temporal sequences of data~\cite{shumway2000time}, and is among the most ubiquitous  mathematical methods. Indeed, time-series are universal for signal processing methods and in pattern recognition applications, dominating characterization of econometrics and finance along with almost any scientific and engineering application.  Time-series methods can be broadly divided into time-domain and frequency-domain methods, the former of which uses a variety of statistical techniques to characterize a sequence, and the latter of which uses spectral (e.g. Fourier) decompositions and wavelets as the underlying representation of the signal.  While the goals of time-series analysis are diverse, including signal estimation, signal classification, signal segmentation, and prediction and forecasting, time-series analysis methods almost universally aim to construct a simple and accurate model underlying the observations.
Noisy, non-stationary data with anomalous exogenous events are the most difficult to characterize for any method.  However, emerging machine learning methods, which are advocated here, allow for a flexible mathematical framework capable of characterizing even such difficult time-series.  We leverage recent advances in long-term probabilistic forecasting, namely the {\em Deep Probabilistic Koopman} (DPK) method~\cite{lange2021fourier,mallen2021deep}, to build a general method for classifying anomalies in multi-dimensional time-series data.  DPK combines deep learning with a spectral estimation approach in order to build a model that accommodates noisy, nonstationary time-series data, further allowing for the detection of anomalous events.


The importance of time-series analysis has led to a broad range of mathematical innovations characterizing sequential temporal data.  In addition to traditional statistical methods, there are growing number of methods using concepts from dynamical systems~\cite{hoffmann2021deeptime,Kutz2016book,Brunton2019book} and machine learning~\cite{dietterich2002machine,masini2021machine} for the analysis of time-series data.
Traditional statistical methods include the time-domain methods, such as the family of {\em autoregressive} (AR) models and their many variants, including ARMA (AR moving average), ARIMA (AR integrated moving average), SARIMA (seasonal ARIMA), etc.~\cite{shumway2000time}.    Such models use a diversity of optimization techniques to estimate parameters of a linear model with its history dependence.  Traditional frequency-domain methods use the properties of short-time Fourier transforms~\cite{kutz2013data} and/or wavelet transforms~\cite{mallat1999wavelet} in order to characterize the signal in a joint time-frequency representation.  More recently, there have been efforts to model time-series data as from a dynamical systems perspective~\cite{hoffmann2021deeptime,Kutz2016book,Brunton2019book}.  Thus the basic concept is to identify a dynamical system as a generative model that underlies the source of the temporal sequence of data.  Included in such methods are the {\em dynamic mode decomposition}~\cite{schmid2010dynamic,Rowley2009jfm,Kutz2016book,Askham2018siads}, Koopman operator~\cite{koopman1931hamiltonian,mezic2005spectral,brunton2021modern,budivsic2012applied,mezic2013analysis,arbabi2017ergodic} and SINDy (sparse identification of nonlinear dynamics)~\cite{brunton2016sindy,Champion2019pnas,Brunton2019book}.  A third approach is offered by machine learning, whose flexible frameworks and universal approximation properties~\cite{hornik1991approximation}, have led to a diversity of neural network architectures~\cite{goodfellow2016deep,yu2019review} for learning time-sequences, including recurrent neural networks (RNNs) and long short-term memories (LSTMs)~\cite{hochreiter1997long}. Although often lacking in interpretability and generalization, deep learning methods have typically exhibited superior performance when provided sufficient training data.

\begin{figure*}[t]
    \centering
    \includegraphics[width=0.85\linewidth]{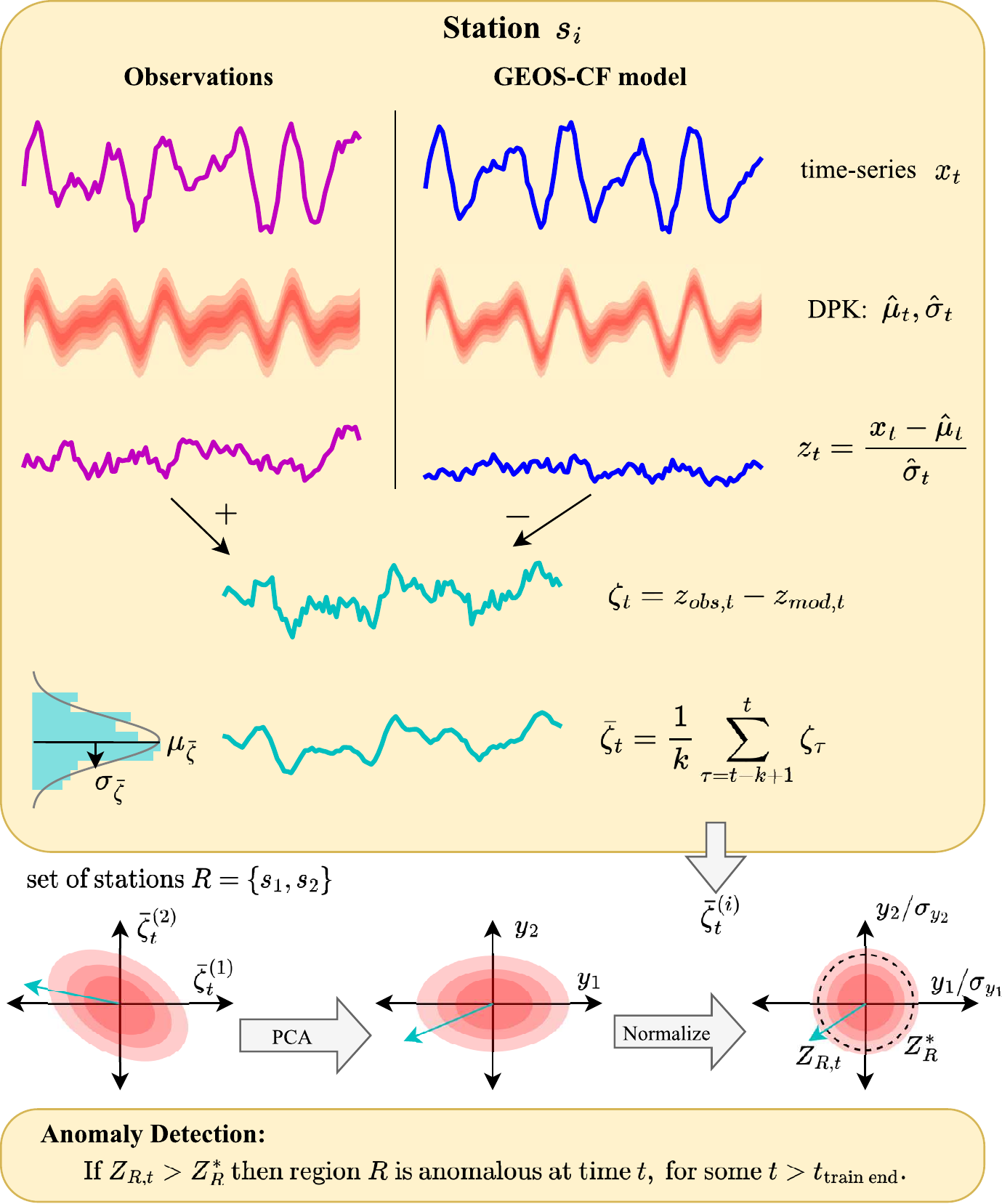}
    \caption{Summary of our method as applied on atmospheric data with Gaussian likelihood. The process shown in the yellow box is performed on each atmospheric monitoring station \(i\) over the training interval: DPK is run on the GEOS-CF model and observations to create a probabilistic “business as usual” model of both, the deviation \(z_t\) from usual is calculated, the difference between these deviations is calculated, and a running average over \(k\) hours is taken to produce \(\bar \zeta_t^{(i)}\), which is approximately normally distributed. If we wish to check whether there is an anomaly in some region \(R = {i_1, …, i_n}\) at some time \(t\) after the training interval, we check whether the Mahalanobis distance \(Z_{R, t}\) of \((\bar \zeta_t^(i_1),\dots, \bar \zeta_t^(i_n))^T\) is larger than the critical value \(Z_R^*\).}
    \label{fig:method}
\end{figure*}

\begin{figure*}[t]
    \centering
    \includegraphics[width=1.05\linewidth]{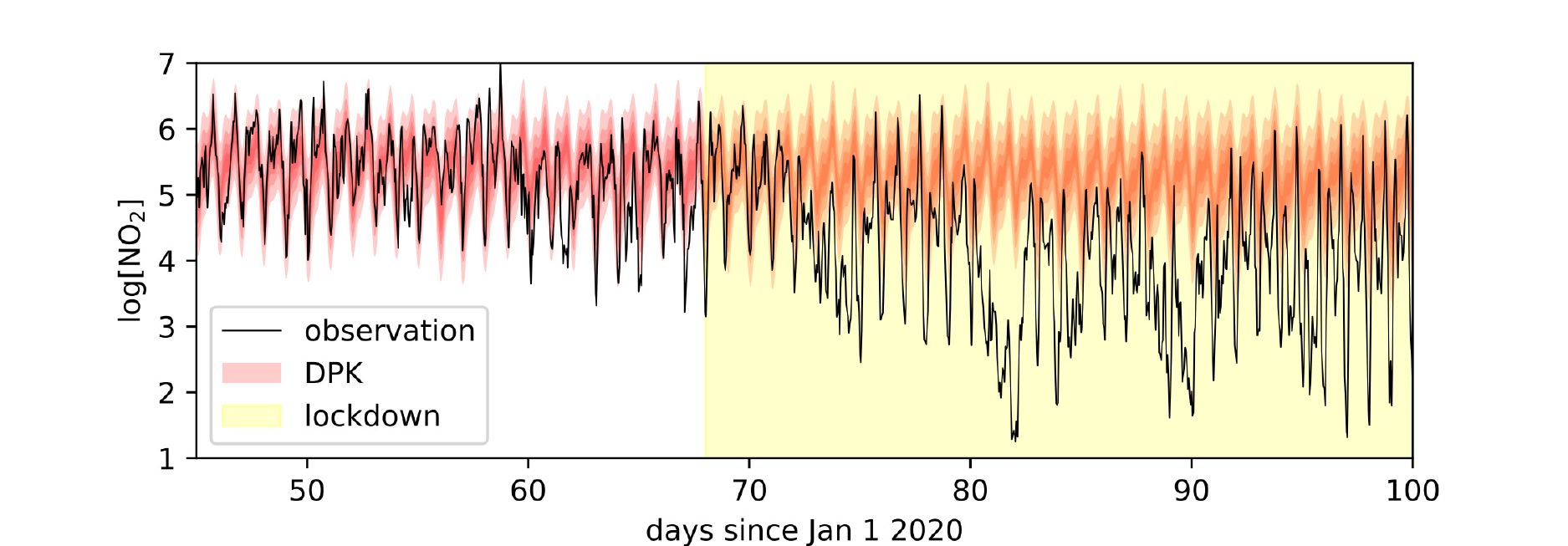}
    \caption{COVID-19 lockdown causes a divergence from the usual atmospheric NO$_2$ concentrations in Milan, Italy. DPK produces a probabilistic forecast of a time-series based on quasiperiodic patterns. The red regions represent the 20\%, 40\%, 60\%, and 80\% confidence intervals of the NO$_2$ concentration according to DPK. Lockdown began in Italy around March 9, 2020 (day 68).}
    \label{fig:madrid_DPK}
\end{figure*}
%
Here we present a novel and general method for classifying anomalies from multi-dimensional time-series data which leverages recent advances in long-term probabilistic forecasting, namely DPK.  DPK is a deep learning framework which takes time-series data and infers a probabilistic model whose parameters are expressible as a function of a minimal set of sinusoids in time~\cite{mallen2021deep}. This fitting procedure captures both the nonstationary aspects of the time-series data, as well as a Fourier-like Koopman decomposition of the signal in time. To demonstrate the capabilities of this method, we apply it to time-series of nitrogen dioxide (NO$_{2}$), an important air pollutant closely linked to the burning of fossil fuel. The evolution of NO$_{2}$ at a given location is fundamentally non-stationary in nature with unknown exogenous influences.  Finding outliers in nonstationary time-series data is a challenging task because it involves creating a long-term probabilistic model of the data that generalizes to future timesteps. Unlike most other machine learning problems which are interpolative (i.e., the inference-time data usually lies in the convex hull of the training data), this problem is extrapolative, and requires training on data from  thousands of timesteps prior to inference for sufficient data size and statistical power. The presented method is able to characterize air pollution anomalies, such as changes in NO$_{2}$ in the wake of COVID-19 lockdowns. Moreover, the method detects these changes only few days after the onset of the anomaly, which makes it suitable for near real-time monitoring applications. An important aspect of our demonstration application is the combined use of publicly available atmospheric observations and corresponding computer simulations. Computer models, such as the NASA Goddard Earth Observing System (GEOS) Composition Forecast system (GEOS-CF) \cite{kellerCF2021}, are an indispensable tool to analyze and understand the drivers of air pollution, such as pollution transport, precursor emissions, and atmospheric chemistry~\cite{brasseur_jacob_2017}. We show how leveraging this information with observations can help reduce type I and type II errors. 

\section{METHODOLOGY} \label{methods}
We first present the basic methodology for classifying anomalous behavior in 1-dimensional time-series observations, then we show how this definition can optionally be expanded to leverage domain-specific models of the observations, and finally we present how to classify anomalies in multi-dimensional observations, allowing for more general and nuanced hypothesis testing. A summary of the methods can be seen in figure~\ref{fig:method}, and an example is shown in figure~\ref{fig:madrid_DPK}.

\begin{figure*}[t]
    \centering
    \begin{overpic}[width=0.95\linewidth]{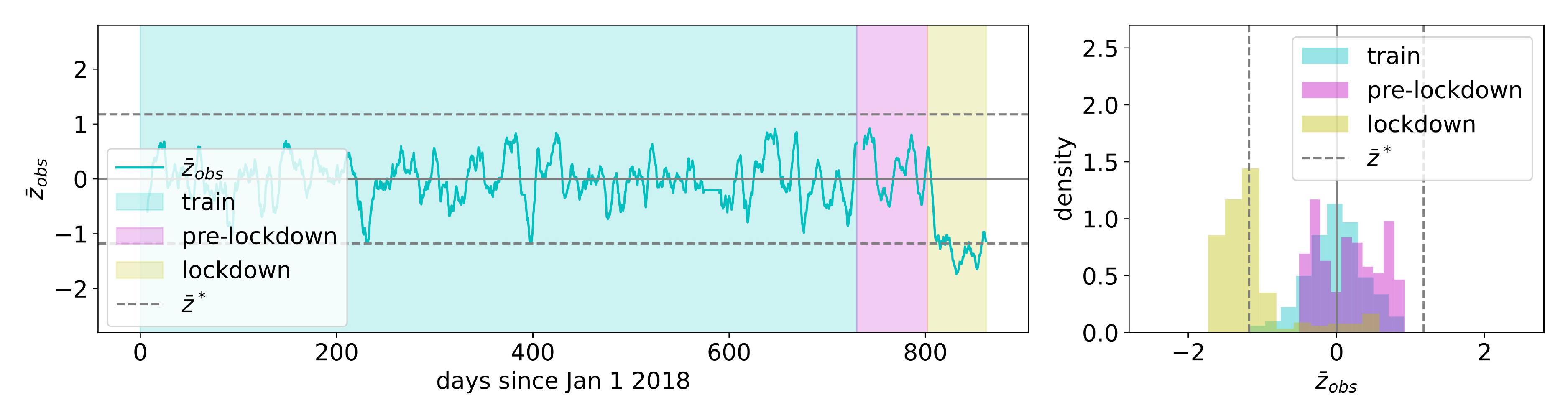}
        \put(0, 24){\Large (a)}
    \end{overpic}
    \begin{overpic}[width=0.95\linewidth]{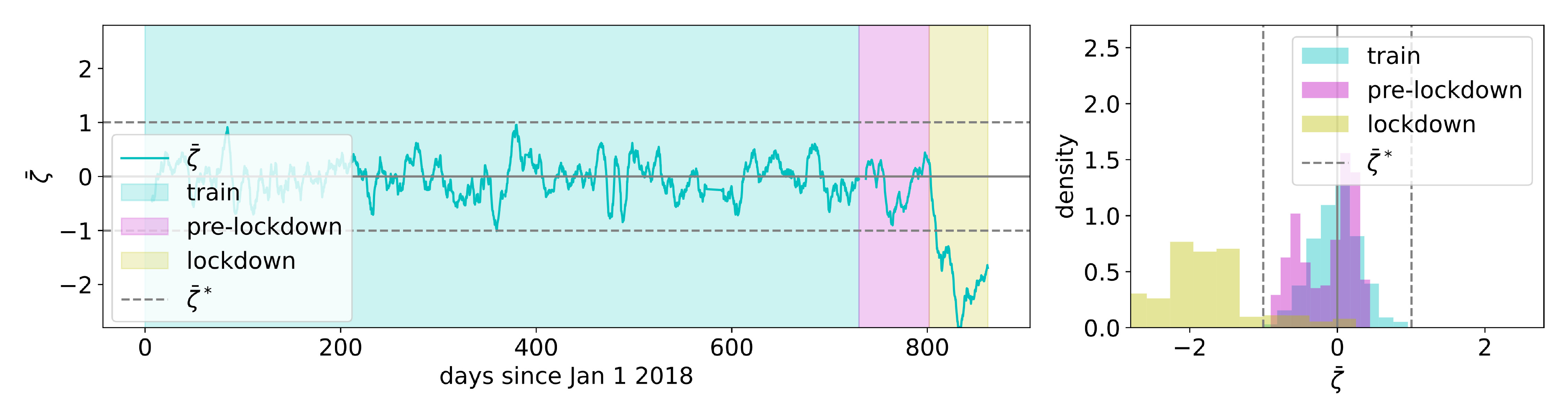}
        \put(0, 24){\Large (b)}
    \end{overpic}
    \caption{Deviation from usual NO$_{2}$ concentration in a monitoring station in Madrid, Spain. After training a DPK model on 2 years of measurements ending on Jan 1 2020, we observe that the concentration of NO$_{2}$ becomes significantly lower during COVID-19 lockdown than it usually is at that time of year according to DPK. \textbf{(a)} (\(\bar z_t\) time-series) This observation can be seen in the sharp decrease in mean z-score, which measures the standardized deviation from typical behavior as predicted by DPK. \textbf{(b)} (\(\bar \zeta_t\) time-series) This deviation persists even when accounting for business-as-usual factors, such as weather fluctuations, as derived from the GEOS-CF model. Critical values are calculated at the \(10^{-3}\) alpha level, and \(k=168\).}
    \label{fig:z_vs_zeta}
\end{figure*}

\subsection{DPK for Anomaly Detection} \label{methods1}
Here we present how to leverage the long-term probabilistic forecasting ability of DPK to produce a calibrated anomaly classifier and monitor for such events in 1-dimensional time-series. 

DPK produces a stable long-term probabilistic forecast of time-series data by expressing it as a parametrized probability distribution (in this case Gaussian) whose parameters vary as a function of sinusoids with different frequencies, as well as a limiting zero frequency for aperiodic trends. It is a result of Koopman operator theory~\cite{mezic2005spectral,Rowley2009jfm,budivsic2012applied,mezic2013analysis,Brunton2021koopman} that any deterministic stable dynamical system can be expressed as a function of sinusoids with various frequencies \cite{koopman1931hamiltonian, koopman1932dynamical, lange2021fourier}. This is extended to the case of stochastic stable dynamical systems by letting our mapping from the sinusoidal Koopman observables be stochastic. Mathematically,
\begin{equation}
    X_t \sim P\left(\vec \theta_t = g_\Theta \left(\begin{bmatrix}
                \cos(\Vec{\omega} t)\\ 
                \sin(\Vec{\omega} t)
            \end{bmatrix}\right)\right),
\end{equation}
where \(\vec \omega\) is a vector of frequencies, \(P(\vec \theta_t)\) is a parametrized probability distribution such as \(\mathcal N\), \(\vec \theta_t\) are the distribution parameters, and \(g_\Theta\) is a feedforward neural network parametrized by \(\Theta\). In this model \(x_t\) is drawn from \(X_t\) for all \(t\). The neural network parameters \(\Theta\) are trained through maximum likelihood. DPK can be understood as providing a ``business as usual" forecast of a time-series based on seasonal patterns.

\begin{figure}[t]
    \centering
    \includegraphics[width=\linewidth]{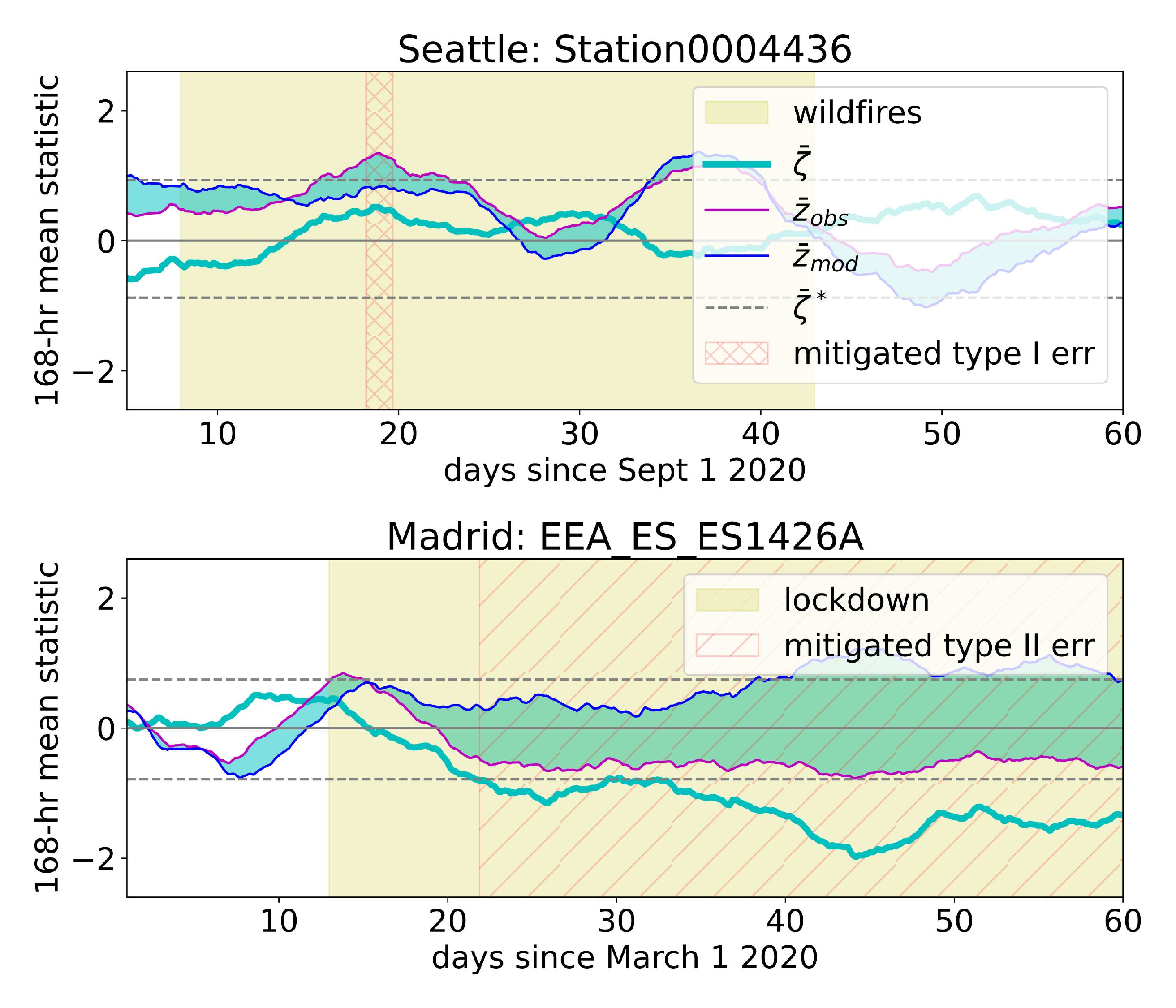}
    \caption{Leveraging a domain-specific model to reduce type I and type II error. When using the \(\bar \zeta\) time-series, where \(\bar \zeta = \bar z_{obs} - \bar z_{mod}\), observations only become significant when they deviate from usual significantly differently than the domain-specific model deviates from usual. We compare two circumstances: one in Seattle where there is a wildfire that is accounted for by the GEOS-CF model (type I error reduction), and another in Madrid where COVID-19 lockdown is not accounted for by the model, which actually expects unusually high NO$_{2}$ due to weather circumstances (type II error reduction). Notice that in Seattle, although the observed NO$_{2}$ in magenta deviates significantly from usual, it is closely tracked by the model NO$_{2}$ in blue, so the \(\bar \zeta\) value remains insignificant. This can be understood as the GEOS-CF model explaining away the deviations in the observations, in this case due to a series of known wildfires. Meanwhile in Madrid, the GEOS-CF model predicts higher-than-usual NO$_{2}$ concentrations due to unusual meteorology, while the observed NO$_{2}$ drops due to COVID-19 lockdown (which GEOS-CF is not aware of), causing the zeta score to drop below the critical value.}
    \label{fig:seattle_v_madrid}
\end{figure}

Once a DPK model has been trained on the time-series of observations, it is used for anomaly detection. First consider the case of univariate time-series for which there is no domain-specific model (such as the GEOS-CF model). Consider the time-series of observed values \(x_{obs, t}\). For a choice of time-varying distribution \(P( \vec \theta_t)\), we define \(z_{obs, t} =\text{ICDF}_{\mathcal N(0, 1)}(\text{CDF}_{P(\vec \theta_t)}(x_{obs, t}))\), where CDF denotes the cumulative distribution function and ICDF denotes the inverse CDF. When \(P\) is a Gaussian this is simply the z-score \((x_{obs, t} - \hat \mu_{obs, t})/\hat \sigma_{obs, t}\). The distribution of \(z_{obs, t}\) is Gaussian because \(\text{CDF}_{P(\vec \theta_t)}(x_{obs, t})\) is uniform when the DPK model is calibrated. Thus, a time is classified as anomalous based on the average of \(z_{obs, t}\) over \(k\) consecutive timesteps, \(\bar z_{obs, t} = \frac{1}{k}\sum_{\tau = t-k+1}^t z_{obs, \tau}\). Consecutive values of \(z_{obs, t}\) are positively correlated due to persistence, so we cannot apply the central limit theorem to obtain the sampling distribution of \(\bar z_{obs, t}\). Instead, the empirical sampling distribution of \(\bar z_{obs, t}\) is used, and fit with a Gaussian distribution \(\mathcal N(\mu_{\bar z_{obs}}, \sigma_{\bar z_{obs}})\). The sampling distribution of \(\bar z_{obs, t}\) is Gaussian because the distribution of \(z_{obs, t}\) is Gaussian. We observe normality empirically, as seen in figure~\ref{fig:z_vs_zeta}a. A time \(t\) is considered anomalous if and only if \(\bar z_{obs, t}\) is as extreme or more extreme than the two-tailed critical value \(\bar z_{obs}^*\) for \(\mathcal N( {\mu_{\bar z_{obs}}}, {\sigma_{\bar z_{obs}}})\) at a certain \(\alpha\) level. These anomalous events represent times in which the recent values of \(x_{obs, t}\) cannot be explained by usual variation as modeled by DPK.

Because the training distribution is supposed to represent observations which are not anomalous, we remove outliers from the training distribution before computing \(\mu_{\bar z_{obs}}\) and \(\sigma_{\bar z_{obs}}\). We discard \(\bar z_{obs}\) values not in \([Q_1 - \lambda \text{IQR}, Q_3 + \lambda \text{IQR}]\) where \(Q_1, Q_3\) are the first and third quartiles, IQR is the interquartile range, and we choose \(\lambda=2\).

It should also be noted that long-term probabilistic forecasting is subject to overconfidence due to imperfect extrapolation and overfitting. DPK is typically not more than 10 to 30\% overconfident, as measured by the root mean square of test-time z-scores, but this overconfidence is accounted for in our selection of the \(\alpha\) level via a hyperparameter sweep in~\ref{hyperparameter}.

\begin{figure}[t]
    \centering
    \includegraphics[width=1\linewidth]{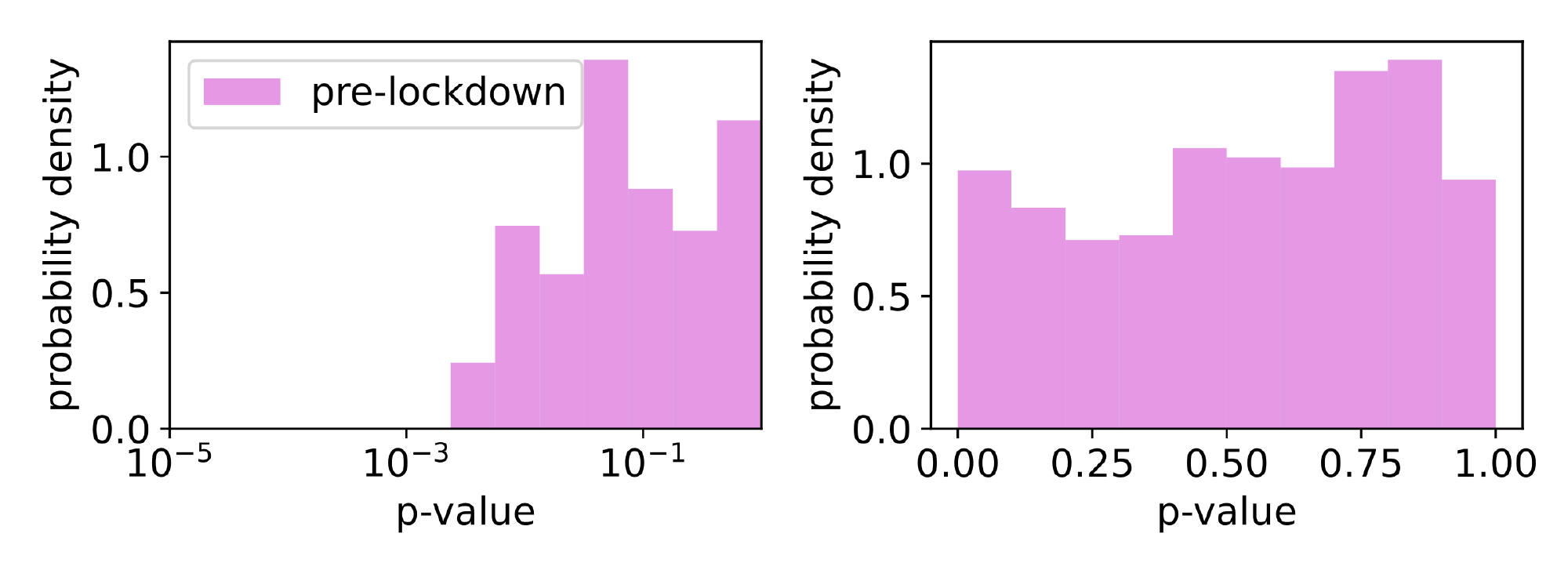}
    \caption{Distribution of p-values pre-lockdown, averaged over several cities, shown in both log and linear scale. During periods without known anomalies, p-values are approximately uniformly distributed, indicating a calibrated method. While figure~\ref{fig:hierarchy} shows that the method is calibrated for univariate time-series, this figure shows that the method is also calibrated for high dimensional time-series (typically a few dozen stations), and does not produce excess false positives.}
    \label{fig:multidim_calibration}
\end{figure}

\begin{figure*}[t]
    \centering
    \begin{overpic}[width=1\linewidth]{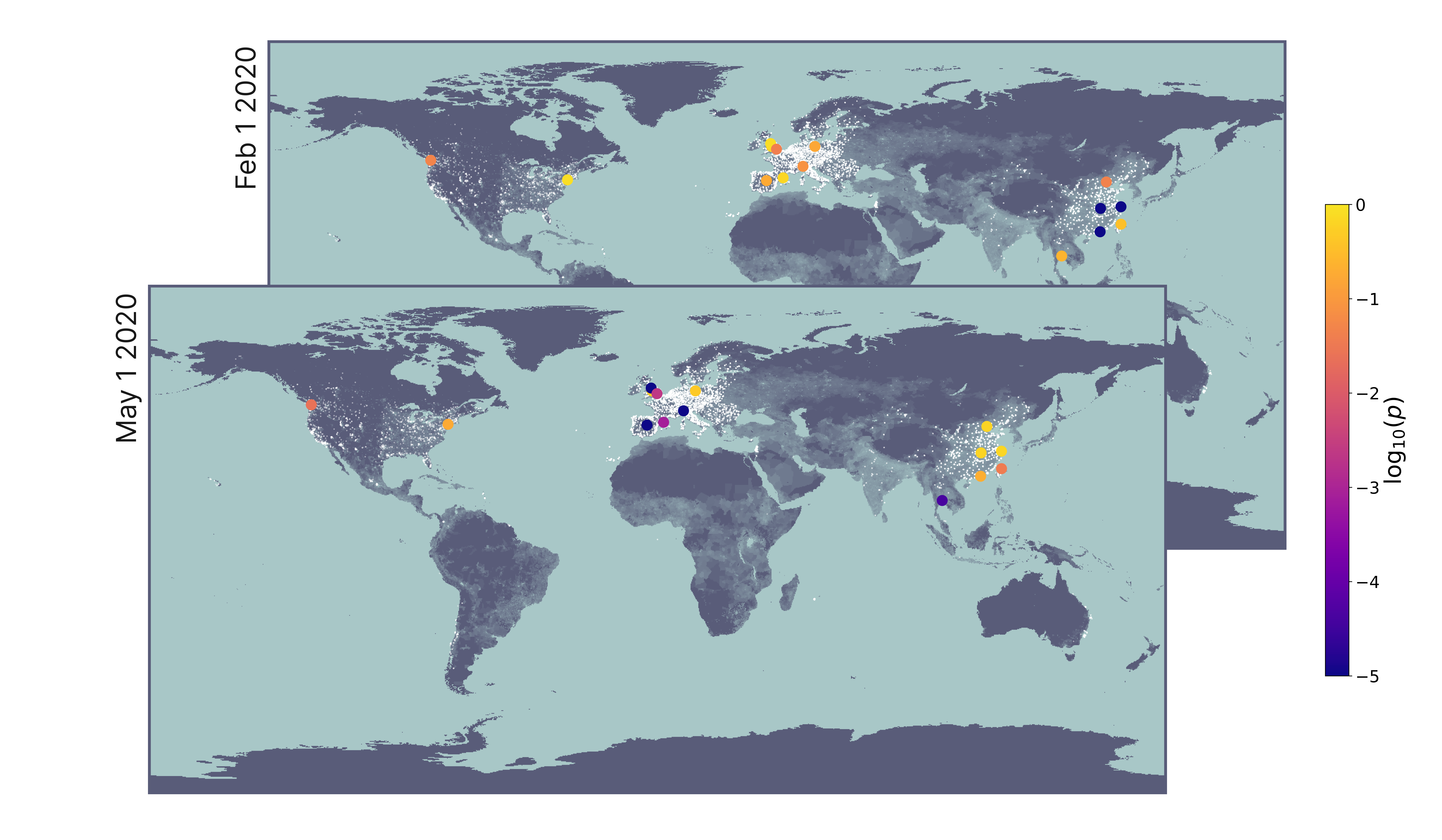}
        \put(0, 54){\Large (a)}
    \end{overpic}
    \begin{overpic}[width=1\linewidth]{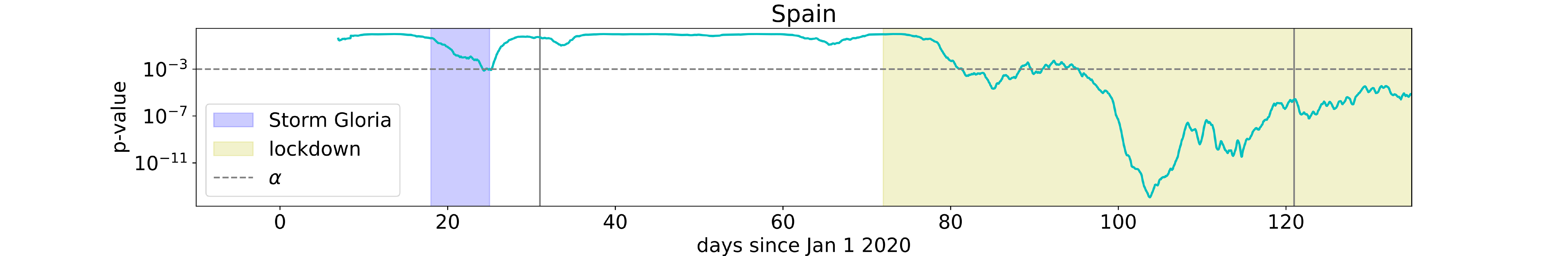}
        \put(0, 14){\Large (b)}
    \end{overpic}
    \begin{overpic}[width=1\linewidth]{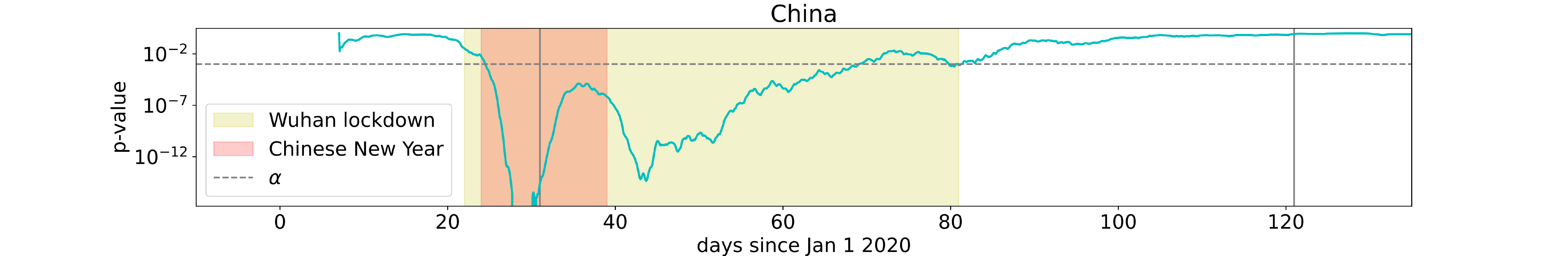}
        \put(0, 14){\Large (c)}
    \end{overpic}
    \caption{A global monitoring system. \textbf{(a)} Of the major cities shown in colored dots, with darker colors indicating more anomalous NO$_2$ values, Chinese cities can be seen experiencing the effects of lockdown first, followed later by cities outside of China. By May 1, Chinese activity is roughly back to normal while European cities and Bangkok still show the effects of lockdown. \textbf{(b-c)} The multivariate anomaly detection method allows for anomalies to be detected at the level of countries. For example, \textbf{(b)} the effects of Storm Gloria and COVID-19 lockdown can be seen in Spanish cities, and \textbf{(c)} the effects of Chinese New Year and COVID-19 lockdown can be seen in Chinese cities.}
    \label{fig:global_snapshots}
\end{figure*}

\subsection{Incorporating domain knowledge}
We can incorporate domain-specific models into our anomaly detection system such that known events are not considered anomalies. This is an optional modification to the technique, and the remaining contributions in section~\ref{methods3} should be of general applicability if this step is skipped because no such model is available for the application. 

When a domain-specific model's predictions \(x_{mod, t}\) are available, we let \(\zeta_t = z_{obs, t} - z_{mod, t}\) and use \(\zeta_t\) in place of \(z_{obs, t}\). \(z_{mod, t}\) is calculated in the same way as \(z_{obs, t}\). As before, an anomaly is flagged when the current \(\bar \zeta_t\) is extreme in the sampling distribution \(\mathcal N(\mu_{\bar \zeta}, \sigma_{\bar \zeta})\) of \(\bar \zeta_t\). Figure~\ref{fig:z_vs_zeta} shows a comparison between using \(\bar z_{obs}\) and using \(\bar \zeta\).

\subsection{Extending to multidimensional time-series} \label{methods3}
Given a set of observables \(R = \{s_1, ..., s_n\}\), we wish to determine whether the entire system those observables measure is anomalous, as shown in figure~\ref{fig:global_snapshots}. Let \(\bar \zeta_{R, t} = (\bar \zeta_t^{(s_1)}, \dots, \bar \zeta_t^{(s_n)})^T\), where \(\bar \zeta_t^{(s_i)}\) is the \(\bar \zeta_t\) described in section~\ref{methods2} for station \(s_i\). Let our multivariate Gaussian estimate of the distribution from which \(\bar \zeta_{R, t}\) is drawn be denoted \(\mathcal N(\mathbf {\mu}_R, {\Sigma}_R)\).

Assuming a multivariate normal distribution allows us to bypass the curse of dimensionality and estimate the joint density using the product of the marginals in PCA space. While PCA produces linearly independent random variables, it does not guarantee statistically independent random variables. However, in most applications it seems reasonable to assume that any statistical dependence between the residual errors of different observables is approximately linear, such that PCA produces approximately statistically independent random variables. This is can be supported by visually verifying random pairs of PCA RVs are independent (granted, pairwise independence does not imply mutual independence). We empirical verify of this method's calibration (see section~\ref{results}) as a test of all assumptions.

The Mahalanobis distance between an observation \(\mathbf d \in \mathbb R^n\) and a distribution \(D\) with mean \(\mathbf {\mu_d}\) and covariance matrix \(S_\mathbf d\) is given by \begin{equation}
    d_M(\mathbf d, D) = \sqrt{(\mathbf d - \mathbf {\mu_d})S_{\mathbf d}^{-1}(\mathbf d - \mathbf {\mu_d})}. \label{eq:mahalanobis}
\end{equation} After performing PCA so that \(\mathbf y = V^T(\mathbf d - \mathbf {\mu_d})\), where the rows of \(V^T\) are the eigenvectors of the empirical covariance matrix, we have \(\mathbf {\mu_y} = \mathbf 0\) and \(S_{\mathbf y} = \text{diag}(\sigma_1^2, ..., \sigma_n^2)\) is a diagonal matrix of the variance along each principal component. Then we have \begin{equation}
    \delta_M(\mathbf d, D) = \left|\left|\begin{bmatrix}y_1 / \sigma_1 \\ \vdots \\ y_n / \sigma_n \end{bmatrix}\right|\right|_2. \label{eq:PCAmahalanobis}
\end{equation}

\begin{figure*}[t]
    \centering
    \begin{overpic}[width=0.95\linewidth]{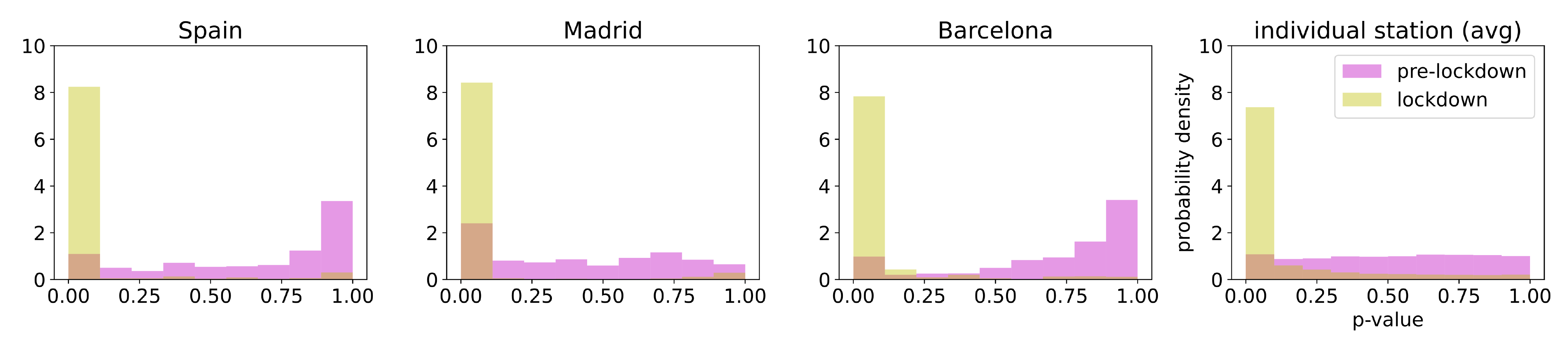}
        \put(-2, 21){\Large (a)}
        \put(23, 21){\Large (b)}
        \put(48, 21){\Large (c)}
        \put(73, 21){\Large (d)}
    \end{overpic}
    \begin{overpic}[width=0.95\linewidth]{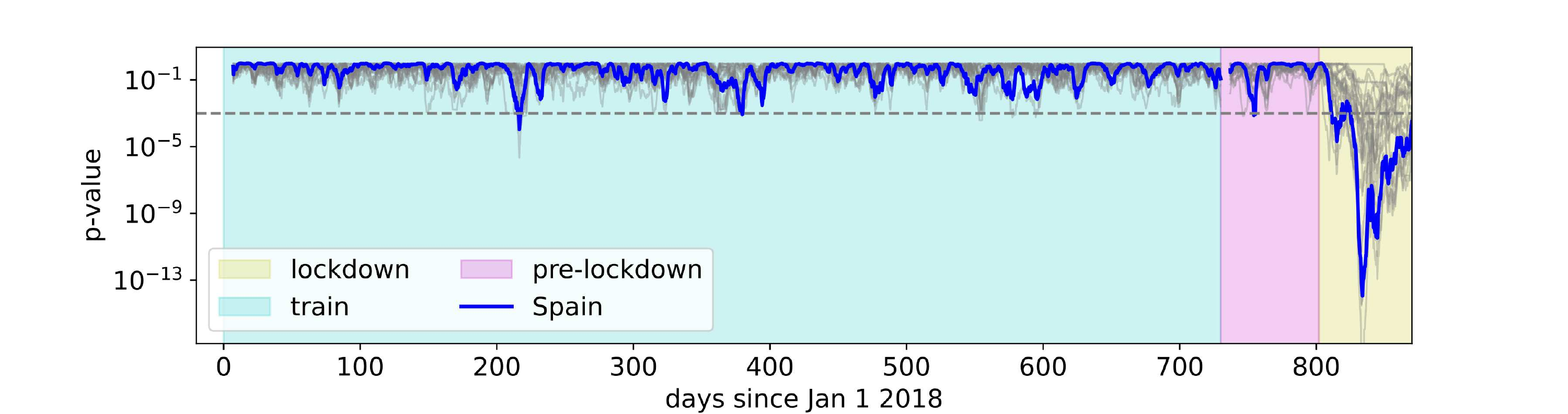}
        \put(3, 24){\Large (e)}
    \end{overpic}
    \caption{Monitoring for anomalies in a hierarchy of stations has multiple advantages: 1) being able to ask a wider variety of questions about NO$_2$ anomalies. 2) being able to leverage more data to evaluate questions. \textbf{(a-d)} The distribution of p-values during and before lockdown at various levels of analysis. \textbf{(e)} The most robust signal is seen in the p-values that leverage the largest amount of data, as can be seen in the blue line representing all of Spain. }
    \label{fig:hierarchy}
\end{figure*}

\(\mathbf d\) is drawn from a multivariate Gaussian, so \((y_1 / \sigma_1, ..., y_n / \sigma_n)^T\) is drawn from a spherical Gaussian with mean \textbf{0} and radius 1, and the p-value of \(\mathbf d\) can be calculated by integrating the probability density over the volume lying outside of the \(n\)-ball of radius \(\delta_M(\mathbf d, D)\) centered at the origin. Let \(Z_{R, t} = \delta_M( \bar{\mathbf \zeta}_{R, t}, \mathcal N(\mathbf {\mu}_R, {\Sigma}_R))\). Then the p-value, the probability of making an observation as extreme or more extreme than a given observation \(Z_R,t\), is given by 
\begin{equation}
    p = P(Z_R \ge Z_{R,t}) \propto \int_{Z_{R, t}}^\infty e^{-Z^2}Z^{n-1} \mathrm{d}Z,
\end{equation}
for \(Z_R,t \ge 0\).

Unlike the usual Mahalanobis distance, we don't use all \(n\) principal components shown in equation~\ref{eq:PCAmahalanobis}, since the components with the smallest variance \(\sigma_i^2\) are mostly noise. Instead, we choose the smallest dimensionality that explains at least 90\% of the variance~\cite{berkooz1993proper,kutz2013data}, although more sophisticated methods for rank truncation exist~\cite{gavish2014optimal,forkman2019hypothesis}. See the appendix figure~\ref{fig:param_sweep} for a hyperparameter sweep demonstrating this.

\section{RESULTS} \label{results}
Here we demonstrate our method using ground-based NO$_{2}$ observations obtained from the OpenAQ platform\footnote{\href{https://openaq.org/}{https://openaq.org/}} and the air quality data portal of the European Environment Agency\footnote{\href{https://discomap.eea.europa.eu/map/fme/AirQualityExport.htm}{https://discomap.eea.europa.eu/map/fme/AirQualityExport.htm}} (EEA). The time-series \(x_{obs, t}\) is the log of the observed concentration.

The DPK model is trained using observations from years 2018 - 2019, and evaluated over 2020. We choose to provide the DPK model with four frequencies \(\omega_i\) observed in our data: daily, weekly, annual, and the long-term trend. We choose \(P\) to be a Gaussian because residuals are approximately normally distributed.

Figure~\ref{fig:madrid_DPK} shows observed NO$_{2}$ concentrations over Milan, Italy (black) for February - May 2020 compared against the corresponding probabilistic DPK forecast (red).  The DPK predicted NO$_{2}$ concentrations agree well with the observations for all of February and the first days of March 2020, but the observations start to diverge from the DPK values afterwards, reflecting the observed decline of NO$_{2}$ in the wake of the COVID-19 pandemic\cite{kellerCovid2021}. By design, this decrease is not captured by DPK, which allows for detection.
 
\subsection{Incorporating the GEOS-CF model} \label{methods2}
The GEOS-CF model is a global atmospheric chemistry model that provides global hourly model estimates of atmospheric composition at $25x25\,\text{km}^{2}$ spatial resolution~\cite{kellerCF2021}, available in near real-time at \footnote{\href{https://gmao.gsfc.nasa.gov/weathe_prediction/GEOS-CF/data_access/}{https://gmao.gsfc.nasa.gov/weather\_prediction/GEOS-CF/data\_access/}}. It is important to note that GEOS-CF uses historical estimates as anthropogenic emissions input, and thus does not capture changes in anthropogenic emission reductions related to COVID-19 restrictions. However, GEOS-CF captures other events that are known to impact air pollution, most notably meteorology and wildfires. 

We only wish to be alerted of anomalies that are not already known by the GEOS-CF model. As can be seen in figure~\ref{fig:seattle_v_madrid}, incorporating GEOS-CF reduces both type I error and type II error, in which an anomaly is veiled by other (e.g., meteorological) factors that push observations in the opposite direction. 

We note that despite having a domain-specific model for this application, GEOS-CF cannot be directly used for anomaly detection because (1) it is not probabilistic, (2) it does not necessarily capture all long-term patterns, and (3) its comparison to observations is complicated by representation errors ($25x25\,\text{km}^{2}$ model averages vs. point source observations) and systematic model biases. DPK is uniquely suited to solve all of these problems because it (1) is probabilistic, (2) solves a global (frequency domain) optimization problem which enables long-term modeling ability, and (3) converges to an unbiased model.

We find the our method successfully detects many known anomalies, and otherwise has a uniform distribution of p-values on the interval [0, 1]. We ran our analysis on various of the largest cities from around the world (filtered for data quality) to validate our method. We chose to look at cities because we expect their air quality to significantly improve when COVID-19 lockdown reduces traffic volume, and because cities have anywhere from a few to several dozen monitoring stations, allowing us to evaluate on a wide variety of time-series dimensionalities. The training interval is from Jan 1 2018 to Jan 1 2020, the explained variance threshold is set to 90\%, and \(k=168\). In figure~\ref{fig:global_snapshots}a the effects of COVID-19 are statistically significant in Chinese cities in February, when the outbreak was mostly limited to China; then by May China has largely resumed normal activity while other major cities are in lockdown, which is reflected in the p-values of NO$_{2}$ concentrations. When looking at the national level in China and Spain (\ref{fig:global_snapshots}b-c), it can be seen that low p-values generally correspond to known exceptional events, including Storm Gloria, COVID-19 lockdown, and Chinese New Year (in which entire cities are largely shut down for days).

Our statistical method is calibrated. In figure~\ref{fig:multidim_calibration}, we see that p-values are roughly uniformly distributed pre-lockdown for various cities. Selection criteria for these cities is described in the appendix. In figure~\ref{fig:hierarchy}d we see that our method is calibrated for individual stations in Spain.

\section{CONCLUSION} \label{conclusion}

We introduced a novel and general method for anomaly detection in complex high-dimensional stochastic time-series data. Our method makes use of recent advances in long-term probabilistic forecasting based on Koopman theory to create a robust ``business as usual" model of observations, against which current observations can be compared. We make use of multidimensional time-series signals to enable online detection of anomalies within several dozen timesteps, or a few days in our application example. The statistical power of the anomaly detection can be improved by incorporating domain-specific numerical models. We demonstrate this using time series of NO$_{2}$ in conjunction with the GEOS-CF numerical atmospheric composition model. Our results show how this technique can be used for near real-time detection of atmospheric anomalies.

When applied to other time-series, such as electricity demand forecasting, we could see our methodological contributions aiding in feature discovery by indicating when observations deviated significantly from the long-term trend. A potential future direction is automatically selecting subsets of the observation vector that might be anomalous, rather than having to check exponentially-many possible subsets or hand-select them.

Given the ubiquity of 
{\em Time-series analysis}, the advocated Koopman based methods provides an powerful  mathematical method for extracting meaningful statistics and characteristics of temporal sequences of data.  Indeed, Koopman theory provides an interpretable framework from which a variety of signal processing goals and pattern recognition applications can be characterized in diverse engineering and scientific fields.  

\section*{Acknowledgements}

JNK acknowledges funding from the National Science Foundation AI Institute in Dynamic Systems grant number 2112085.

\section{APPENDIX}

\subsection{Experimental details}
\textbf{DPK training details.}
The code is available at \href{https://github.com/AlexTMallen/NASA-GEOS}{https://github.com/AlexTMallen/NASA-GEOS}. The DPK model for each station is trained with a learning rate of 0.0001, a weight decay regularization of 0.001, for 400 epochs, on hourly observations from January 1 2018 to January 1 2020. Each model is initialized using the parameters of a model trained on another station's observations to aid efficiency through transfer learning.

\textbf{Selection criteria for cities in Section~\ref{fig:multidim_calibration}}

From a list of major cities around the world, we look at high-pollution (greater than 10 ppb NO$_2$ multi-year average), high-population cities with clean data during the time in 2020 before COVID-19 lockdown. Most cities are excluded because of a lack of clean data, if any is available at all. We also exclude a city if it has a significant p-value that corresponds with a known anomalous event. Here are the cities we excluded due to known anomalies.
Madrid and Barcelona experienced Storm Gloria (\href{https://en.wikipedia.org/wiki/Storm_Gloria}{https://en.wikipedia.org/wiki/Storm\_Gloria}). 
Birmingham, London, Berlin experienced Storm Ciara (\href{https://en.wikipedia.org/wiki/Storm_Ciara}{https://en.wikipedia.org/wiki/Storm\_Ciara}).

\subsection{Hyperparameter selection} \label{hyperparameter}

\begin{figure*}[t]
    \centering
    \includegraphics[width=0.9\linewidth]{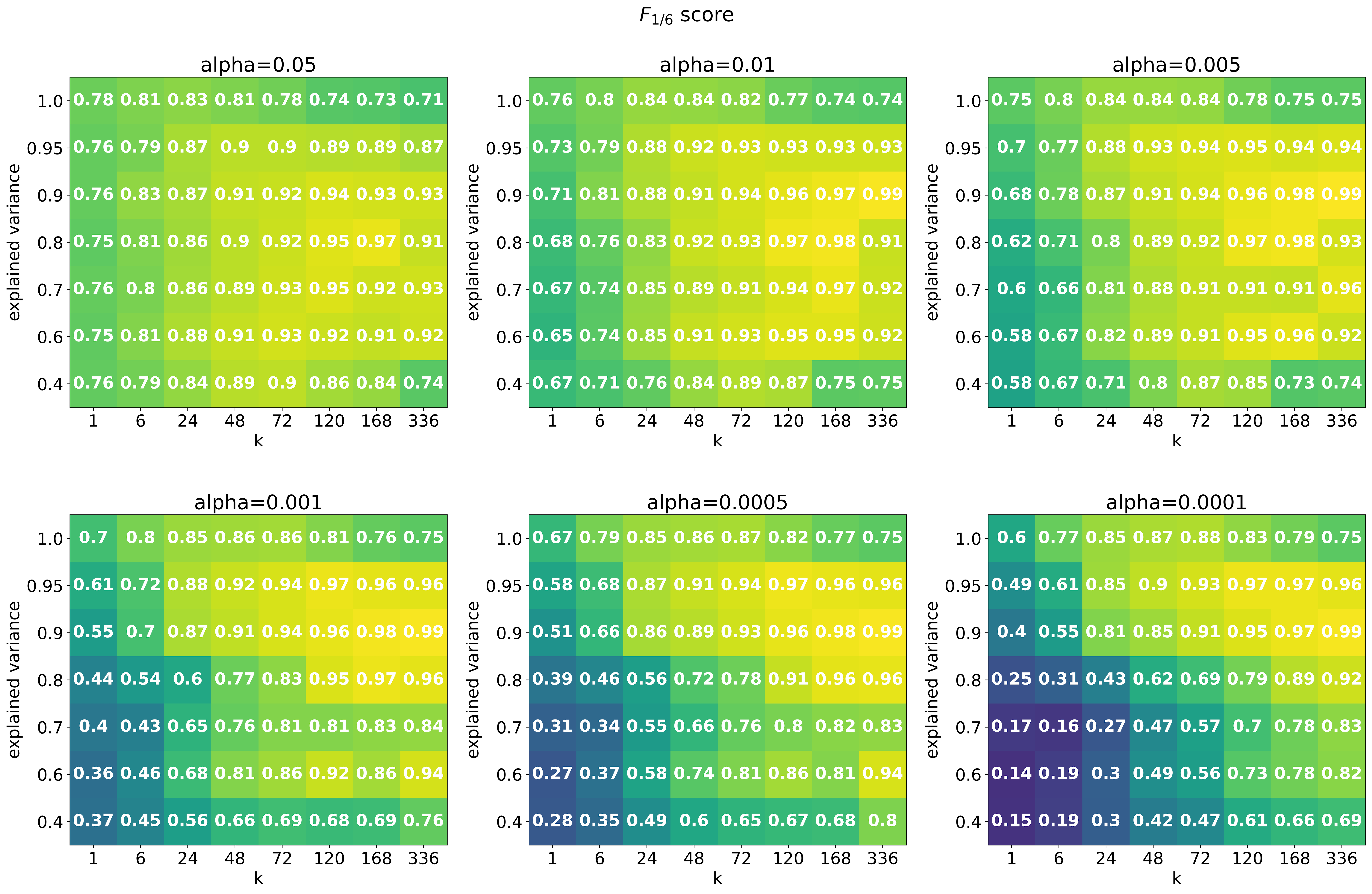}
    \includegraphics[width=0.9\linewidth]{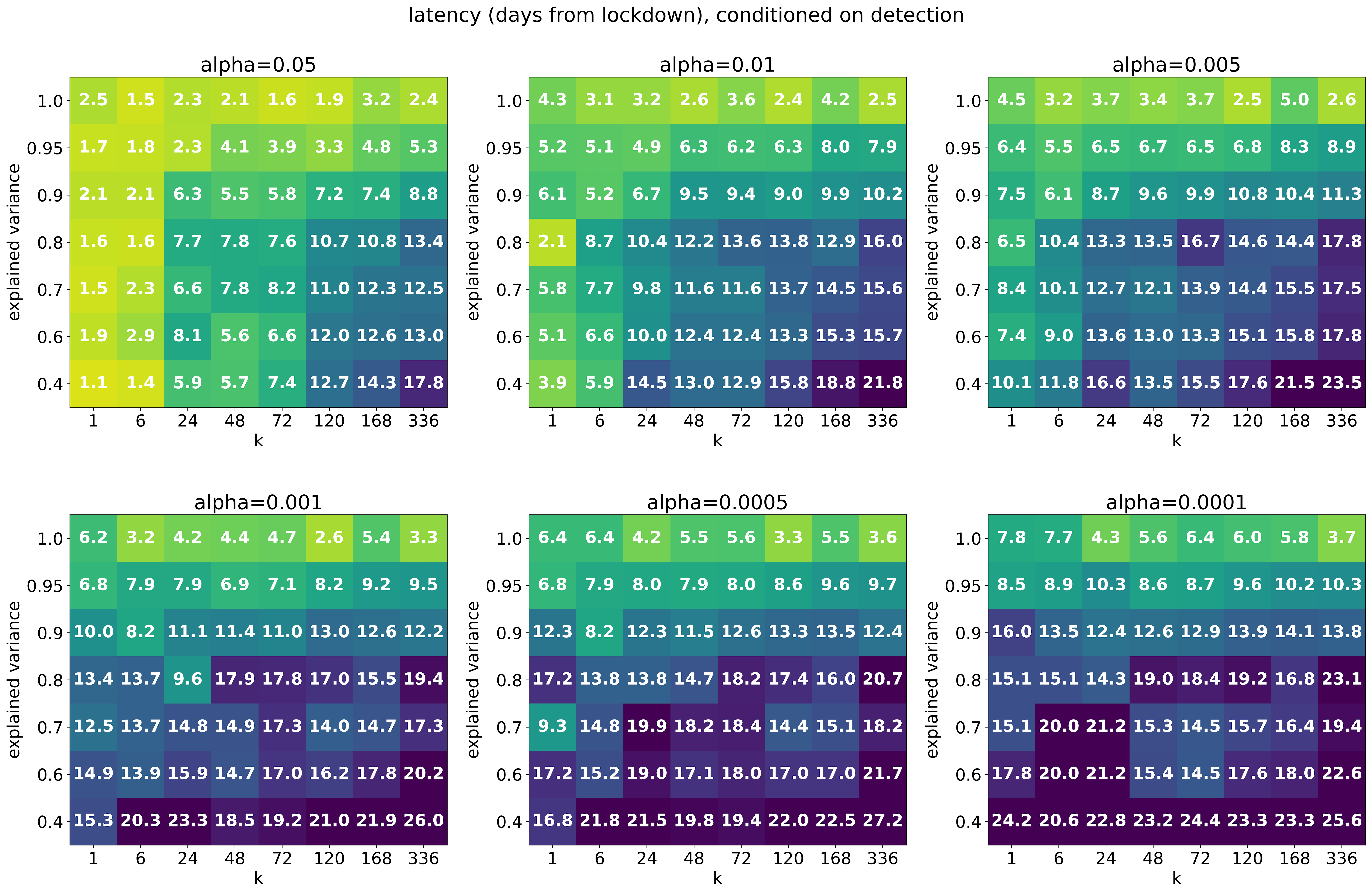}
    \caption{Hyperparameter sweep over the \(\alpha\) level, the number of hours \(k\) over which to average, and the explained variance threshold for calculating \(Z_{R,t}\). We measure both \(F_{1/6}\) classification accuracy and latency of the anomaly detection, conditioned on detection. \(F_{1/6} = (1 + \frac{1}{6^2})(\text{precision} \cdot \text{recall}) / (\frac{1}{6^2}\text{precision} + \text{recall})\) is an accuracy measure that places \(6\) times as much weight on precision as on recall, and we use it because the vast majority of times are not anomalous while only a single time within an anomalous event need be flagged. Based on these results we selected  90\% explained variance, \(k=\)168 hours (1 week) averaging, and \(\alpha= 0.001\).}
    \label{fig:param_sweep}
\end{figure*}

Hyperparameter tuning is critical for almost any data-driven method.  In the application considered here, hyperparameter sweeps are done over the \(\alpha\) level, the number of hours \(k\) over which to average, and the explained variance threshold for calculating \(Z_{R,t}\).  Figure~\ref{fig:param_sweep} summarizes the performance of the hyperparameter sweeps.

\bibliographystyle{unsrt}
\bibliography{main}

\end{document}